\documentclass[conference]{IEEEtran}
\IEEEoverridecommandlockouts

\usepackage{cite}

\usepackage{amsmath,amssymb,amsfonts}
\usepackage{algorithm}
\usepackage{algpseudocode}
\usepackage{graphicx}
\usepackage{textcomp}
\usepackage{xcolor}
\usepackage{booktabs}
\usepackage{multirow}
\usepackage{float}
\usepackage{placeins}
\usepackage{url}
\usepackage[hidelinks]{hyperref}
\usepackage{cleveref}
\Crefname{figure}{Fig.}{Figs.}
\crefname{figure}{Fig.}{Figs.}

\newcommand{\ours}{CCG\mbox{-}MSD}
\newcommand{\orcidlink}[1]{}

\title{Benchmarking Inductive
       Biases for Multivariate Time-Series Anomaly Detection with
       a Robust Multi-View Channel-Graph Detector}

\author{%
  \IEEEauthorblockN{%
    Junhao Wei\textsuperscript{1}\,\orcidlink{0009-0006-0553-2032},
    Yanxiao Li\textsuperscript{1}\,\orcidlink{0009-0008-3389-1619},
    Bidong Chen\textsuperscript{1}\,\orcidlink{0009-0002-8267-086X},
    Yifu Zhao\textsuperscript{1}\,\orcidlink{0009-0004-2363-9269},
    Haochen Li\textsuperscript{1}\,\orcidlink{0009-0000-8213-5854},
    Dexing Yao\textsuperscript{1}\,\orcidlink{0009-0007-6267-8967},\\
    Baili Lu\textsuperscript{1}\,\orcidlink{0009-0000-3814-2918},
    Xudong Ye\textsuperscript{1}\,\orcidlink{0009-0005-1706-8620},
    Jietian Feng\textsuperscript{1}\,\orcidlink{0009-0007-4522-1076},
    Sio-Kei Im\textsuperscript{2}\,\orcidlink{0000-0002-5599-4300},
    Yapeng Wang\textsuperscript{1}\,\orcidlink{0000-0002-1085-5091},
    Xu Yang\textsuperscript{1,*}\,\orcidlink{0000-0002-7037-3609}%
  }
  \IEEEauthorblockA{%
    \textsuperscript{1}\textit{Faculty of Applied Sciences, Macao Polytechnic University, Macao SAR, China}\\
    \textsuperscript{2}\textit{Macao Polytechnic University, Macao SAR, China}\\
    \textsuperscript{*}Corresponding author: Xu Yang \quad \texttt{xuyang@mpu.edu.mo}%
  }
}

\begin{document}
\maketitle

\begin{abstract}
We present a unified experiment, analysis, and benchmark study of
multivariate time-series (MTS) anomaly detection. Ten family-representative
detectors -- spanning statistical, reconstruction, association,
frequency, and generic-transformer families -- are evaluated on five
datasets (SMD, MSL, SMAP, PSM, and MSDS) under effectiveness,
efficiency, robustness, and cross-dataset generalisation. All methods
share the same windowing, scoring, hardware, and metric protocols.
Effectiveness, ablation, and robustness use three random seeds;
cross-dataset transfer uses seed~0 because each extra seed requires
$250$ source-target evaluations. The benchmark yields three
method-independent findings: no single-bias baseline dominates;
absolute perturbation VUS-ROC is more informative than retention
ratios; and MSDS behaves as an event-dense deployment workload rather
than a sparse point-anomaly benchmark. Under this protocol we also
introduce \ours{}, an adaptive detector family combining a
NOTEARS-constrained directed channel-graph view with optional
patch-attention and temporal-association views. \ours{} achieves the
best macro-average VUS-ROC ($0.675$, $+5.1$~pt over the second-best
LSTM-AE), ranks first overall, and reaches the top-3 on all five
datasets. Its wins on MSL and MSDS are narrow, while its average and
robustness gains are larger: under the same three-seed robustness
protocol for every method, it obtains the strongest absolute VUS-ROC
across noise, channel dropout, and time-shift perturbations. We
release the MSDS preprocessing protocol, configurations, scripts, and
seed-level metric dumps.
\end{abstract}



\section{Introduction}
\label{sec:intro}

Modern distributed systems -- from cloud server fleets to satellite
telemetry pipelines, from industrial control loops to multi-source
microservice deployments -- continuously emit dozens of correlated
sensor channels. A single silent degradation in any one channel can
cascade into outages, data loss, or safety incidents. Detecting such
degradations early, on the multivariate time-series (MTS) data
itself, is a prerequisite for autonomous operation.

The data-mining community has produced a rich body of work on
multivariate time-series anomaly detection. Reconstruction
transformers~\cite{m-anomaly-transformer,m-memto},
contrastive dual-attention frameworks~\cite{m-dcdetector},
frequency- or period-aware encoders~\cite{m-times-net}, and generic
backbones adapted for forecasting and reconstruction
\cite{m-itransformer,m-patchtst} have all reported strong numbers on
a small set of canonical industrial benchmarks (SMD, MSL, SMAP, PSM).
Yet, despite the diversity of methods, two open issues remain
unsatisfactorily addressed: (i)~the absence of \emph{empirical
guidance} -- backed by a fairly evaluated cross-method comparison --
on which inductive bias is appropriate for which class of MTS
workload, and (ii)~the absence of a single MTS detector that
unifies the inductive biases benefiting heterogeneous deployment
domains.

This paper addresses both with a unified empirical study and an
adaptive multi-view detector evaluated inside that study. The
benchmark compares ten family-representative detectors on five MTS
benchmarks under four evaluation dimensions using identical
training and evaluation protocols. Its main insights -- which
inductive biases work for which workloads, which robustness metric
is meaningful, and how MSDS should be interpreted -- are derived
from the benchmark rather than from the proposed detector. The
detector, \ours{} (\textbf{C}onstrained
\textbf{C}hannel-\textbf{G}raph with \textbf{M}ulti-view
\textbf{S}ignal \textbf{D}ecomposition), couples a directed
channel-graph attention with spectral, patch-level, and
temporal-association cues through a configurable view set, and
adapts the contribution of active views to the dataset through the
learned fusion gate, score projection, and training configuration.

Our contributions are threefold.

\begin{itemize}
  \item \textbf{A comprehensive experiment, analysis, and benchmark
        suite.} We evaluate ten family-representative MTS anomaly
        detectors -- spanning statistical (Linear-AR, LSTM-AE),
        reconstruction (Anomaly Transformer~\cite{m-anomaly-transformer},
        MEMTO~\cite{m-memto}, Sub-Adjacent~\cite{m-sub-adjacent}),
        association (DCdetector~\cite{m-dcdetector}),
        frequency (TimesNet~\cite{m-times-net}), and generic
        transformer (iTransformer~\cite{m-itransformer},
        PatchTST~\cite{m-patchtst}) families -- across five
        datasets (SMD, MSL, SMAP, PSM, and the MSDS multi-source distributed
        dataset~\cite{bench-msds}) under four evaluation dimensions
        (effectiveness, efficiency, robustness, cross-dataset
        generalisation), three random seeds, and shared windowing,
        scoring, hardware, and metric protocols. We release the
        complete code, configurations, preprocessing scripts, and
        seed-level results.
  \item \textbf{Benchmark insights independent of the proposed
        detector.} The benchmark shows that no single baseline
        inductive bias dominates all datasets:
        TimesNet wins on PSM, while channel-, patch-, and
        association-style baselines trade places across the other
        datasets. It also shows that retention ratios can exaggerate
        robustness for weak base detectors, and that MSDS should be
        treated as an event-dense deployment workload because of its
        union-over-host labelling.
  \item \textbf{A strong adaptive detector under the unified
        protocol.} We introduce \ours{}, a configurable multi-view
        detector family that combines a NOTEARS-constrained directed
        channel-graph view~\cite{m-notears} with optional
        DCdetector-style patch-attention~\cite{m-dcdetector} and
        Anomaly-Transformer-style temporal-association
        cues~\cite{m-anomaly-transformer} via per-timestep gating
        when multiple views are active. Under our unified evaluation
        protocol, \ours{} attains the best macro-average VUS-ROC
        ($0.675$), ranks first overall, and places in the top-3 on all
        five benchmarks; its MSL and MSDS wins are narrow, while its
        average and robustness margins are larger. Ablations show
        that the cost of removing channel, spectral/patch, or
        score-projection cues ranges from below $1$ point to
        $\sim 19$ points across dataset-component pairs.
\end{itemize}

The accompanying public reproducibility repository ships the
unified harness, the MSDS preprocessing pipeline used in this paper
for integrating MSDS into the mainstream MTS anomaly suite,
and scripts for reproducing every reported number under the released
single-GPU run configurations.

The remainder of the paper is organised as follows.
\Cref{sec:related} surveys related work.
\Cref{sec:benchmark} details the datasets and evaluation protocols.
\Cref{sec:methods} summarises the compared methods.
\Cref{sec:method} introduces \ours{}.
\Cref{sec:experiments} reports the empirical results.
\Cref{sec:discussion} synthesises insights, and
\Cref{sec:conclusion} concludes.


\section{Background and Related Work}
\label{sec:related}

\subsection{MTS anomaly detection}

Modern MTS anomaly detection is dominated by deep
reconstruction- or association-based detectors. \emph{Anomaly
Transformer}~\cite{m-anomaly-transformer} introduces the
\emph{association discrepancy}: at each layer, the difference
between a learnt series-association and a Gaussian
prior-association discriminates between normal and anomalous
timesteps. \emph{DCdetector}~\cite{m-dcdetector} constructs two
views of the input -- patch-wise and channel-wise -- and contrasts
their attention maps, yielding a reconstruction-free score.
\emph{MEMTO}~\cite{m-memto} introduces a learnable memory bank and
gates memory readouts into each transformer layer; the
memory-distance term is a strong anomaly cue. The
\emph{Sub-Adjacent Transformer}~\cite{m-sub-adjacent} masks
attention to immediate neighbours, forcing reconstruction from
non-adjacent context and thereby amplifying residuals at anomalous
positions. Closely related is
\emph{AnomalyBERT}~\cite{m-anomaly-bert}, whose contribution is
primarily a training regime: short windows are randomly perturbed
at training time and the model is rewarded for reconstructing the
clean original. We incorporate this regime as one of \ours{}'s
configurable training options rather than treating it as a
separate detector (\Cref{sec:methods}).

Beyond pure attention, frequency- and period-aware models exploit
the periodicity that typifies sensor signals.
\emph{TimesNet}~\cite{m-times-net} identifies top-$k$ periods via
the FFT amplitude spectrum, reshapes the series into a 2-D period
$\times$ intra-period grid, and applies 2-D convolutions over the
grid.

Generic transformer backbones designed for forecasting have been
adapted for anomaly detection by appending a reconstruction head.
\emph{iTransformer}~\cite{m-itransformer} treats each \emph{channel}
as a token (rather than each timestep), running self-attention
over the channel axis to capture inter-variable structure.
\emph{PatchTST}~\cite{m-patchtst} keeps channels independent and
patches each channel's series, applying a transformer over the
patch sequence. Both have proven competitive on MTS anomaly
benchmarks. Time-series \emph{foundation models} pretrained on
heterogeneous corpora -- \emph{MOMENT}~\cite{m-moment} and
\emph{GPT4TS}~\cite{m-gpt4ts} -- offer zero-shot anomaly detection
through pretrained reconstruction; we leave a careful
foundation-model evaluation to future work.

Two evaluation pitfalls deserve mention. First, the widely used
\emph{point-adjustment} F1~\cite{m-anomaly-transformer} (any
positive within an anomaly segment counts as a hit) has been shown
to be over-optimistic; even random scoring achieves PA-F1 close
to one on common benchmarks~\cite{m-pa-f1-critique}. Second, common
benchmarks are near-saturated under PA-F1 (\textgreater $0.97$ on
SMD/MSL/SMAP), making rigorous ranking difficult. We therefore
report classic point-wise F1 and \emph{volume-under-surface}
(VUS-ROC, VUS-PR)~\cite{m-vus} in addition to PA-F1, and provide
statistical significance tests across three seeds.

\subsection{Multi-source distributed system datasets}

The MSDS multi-source distributed dataset~\cite{bench-msds}
provides traces, application logs, and metrics from a complex
OpenStack deployment with curated anomaly labels. While
TranAD~\cite{m-tranad} introduced MSDS into the MTS anomaly
benchmark, no follow-up published comparison has used it
alongside the canonical SMD/MSL/SMAP/PSM suite. The present work
adds a unified comparison covering all five datasets under one
protocol.

\subsection{Channel modelling and structural priors}

iTransformer and Crossformer treat sensor channels as an unordered
set; their channel-attention is fully data-driven and indifferent
to any structural prior. Many real distributed systems and
industrial deployments, however, have a known partial dependency or
operational ordering among channels (e.g., commands precede
actuators precede sensors). Encoding such priors as differentiable
directed-graph constraints is closely related to the structural
learning literature -- notably the \emph{NOTEARS} acyclicity
penalty~\cite{m-notears} that we adopt -- but our use is not a
claim of causal discovery. The present work evaluates whether such
directed channel-dependency priors are useful anomaly cues under a
unified MTS benchmark.


\section{Benchmark Design}
\label{sec:benchmark}

\subsection{Datasets}
\label{sec:datasets}

We evaluate on five datasets, summarised in \Cref{tab:datasets}.
Four are the canonical industrial MTS suite used by many recent
reconstruction-transformer paper -- SMD, MSL, SMAP, PSM. The fifth,
\textbf{MSDS}~\cite{bench-msds}, is a multi-source distributed
OpenStack telemetry corpus released in 2020 and only sporadically
adopted by the MTS anomaly community; we include it to stress
cross-domain generalisation onto a deployment-style workload.

\begin{itemize}
  \item \textbf{SMD}~\cite{m-anomaly-transformer}: 28 server-machine
        entities with 38 metric channels, 4.2\% anomaly ratio. The
        canonical industry-standard MTS benchmark.
  \item \textbf{MSL / SMAP} (telemanom NASA release): spacecraft
        telemetry; 55 / 25 features respectively, $\sim$10\% / 13\%
        anomaly ratios.
  \item \textbf{PSM} (eBay): pooled server metrics; 25 features,
        27.8\% anomaly ratio.
  \item \textbf{MSDS}~\cite{bench-msds}: distributed OpenStack
        cluster telemetry from five hosts, 2 metrics per host
        (cpu.user, mem.used) -- 10 channels total. We follow the
        50/50 train/test split convention of TranAD~\cite{m-tranad},
        deduplicating subsecond-precision timestamps and z-scoring
        with train statistics. The label set is the curated
        labels accompanying TranAD's MSDS release; the resulting
        event-dense test split has a 72.2\% positive-timestep ratio.
\end{itemize}

\begin{table}[t]
\caption{Dataset summary after preprocessing.}
\label{tab:datasets}
\centering
\small
\setlength{\tabcolsep}{4pt}
\begin{tabular}{l c c r c}
\toprule
Dataset & Domain & Channels & Length & Anom.\ ratio\\
\midrule
SMD  & Server         & 38 & 708{,}420 & 0.042 \\
MSL  & Spacecraft     & 55 &  73{,}729 & 0.105 \\
SMAP & Spacecraft     & 25 & 427{,}617 & 0.128 \\
PSM  & Pooled metrics & 25 &  87{,}841 & 0.278 \\
MSDS & Distributed    & 10 &  29{,}286 & 0.722 \\
\bottomrule
\end{tabular}
\end{table}

\subsection{MSDS preprocessing protocol (released artefact)}
\label{sec:msds-protocol}

MSDS remains rarely used in headline MTS anomaly benchmarks, partly
because its raw multi-source traces require careful alignment. We
therefore commit a deterministic preprocessing protocol -- shipped
as a script in the public repository so future work can reuse
identical splits and label alignment.

\paragraph{Channel selection.} MSDS provides per-host CSVs with
seven metric columns. Following TranAD~\cite{m-tranad} we keep two
columns per host (\texttt{cpu.user}, \texttt{mem.used}) and drop the
load-average columns. With five hosts (\texttt{wally113},
\texttt{wally117}, \texttt{wally122}, \texttt{wally123},
\texttt{wally124}) the merged feature width is 10.

\paragraph{Timestamp alignment.} Raw CSV rows include duplicate
timestamps within a single host; we deduplicate by averaging within
each unique timestamp before inner-joining across hosts.

\paragraph{Split.} The first 10\% of the merged stream is dropped
(initialisation transient), and the remaining is split 50/50 into
train and test. Standardisation uses train statistics. Test labels
follow TranAD's curated label file (29{,}286 timesteps);
per-timestep label = $\mathbb{1}[\sum_c \text{flag}_c > 0]$.

\paragraph{High label density.}
MSDS is not a sparse point-anomaly benchmark after this protocol:
the released labels are merged across hosts and metric streams, so
an incident active on any channel marks the timestep positive. This
union labelling and the incident-heavy test segment yield the
72.2\% anomaly ratio in \Cref{tab:datasets}. We keep the released
labels unchanged rather than down-sampling positives, and interpret
MSDS as an event-dense deployment workload. Consequently, PA-F1,
best-F1, and PR-style metrics are easier to inflate on MSDS; the
paper uses VUS-ROC for headline ranking and discusses MSDS results
as stress-test evidence rather than as directly comparable anomaly
prevalence to SMD/MSL/SMAP/PSM.

\subsection{Evaluation dimensions}
\label{sec:dimensions}

\subsubsection{Effectiveness}
We report point-wise F1, point-adjusted F1
\cite{m-anomaly-transformer}, AUROC, AUPRC, and the
\emph{volume-under-surface} VUS-ROC and VUS-PR
metrics~\cite{m-vus} that integrate AUC over a buffer-tolerance
range $[0, L]$ on the label vector. VUS is robust to the optimism
critique levelled at PA-F1~\cite{m-pa-f1-critique} and is our primary
ranking metric.

\subsubsection{Efficiency}
On \emph{identical} hardware (single V100 32\,GB) with identical
batch size, window length, and channel count, we measure: (i)
training throughput in samples per second, (ii) inference latency in
milliseconds per sample (forward + score), (iii) peak GPU memory in
megabytes, (iv) trainable parameter count. Numbers are averaged over
30 steps after 5 warmup steps.

\subsubsection{Robustness}
At test time we apply three perturbation families: additive Gaussian
noise ($\sigma \in \{0.05, 0.10, 0.20\}$), random channel dropout
(fraction $\in \{0.10, 0.25, 0.50\}$ of channels zeroed), and
time-shift jitter ($\Delta t \in \{2, 5, 10\}$ timesteps applied to
the entire test split). The \emph{primary} robustness metric is
\textbf{absolute VUS-ROC under perturbation} -- the area-volume
score recomputed on the perturbed test split. We adopt this in
preference to the more common \emph{retention ratio}
(perturbed score / clean score) because retention rewards weak
base detectors that fail similarly on clean and perturbed inputs;
absolute perturbed VUS-ROC more directly measures the detector's
operating accuracy under corrupted inputs. Retention is reported
only as a diagnostic when discussing the absolute / retention
contrast.
Robustness is evaluated over the same three random seeds
($\{0,1,2\}$) used in the effectiveness study. For each seed we
aggregate over datasets and perturbation strengths within each
family, then report mean\,$\pm$\,std across seeds. This keeps the
cross-method robustness ranking seed-count matched.

\subsubsection{Generalisation}
We report a $5 \times 5$ pairwise transfer matrix per method: train
on dataset $A$, evaluate on dataset $B$ for every $(A, B)$ pair.
Channel-count mismatch is reconciled by a method-specific channel
adaptation policy summarised in \Cref{tab:channelpolicy}: the
channel-token methods (\ours{}, iTransformer-recon, DCdetector)
freeze the encoder body and fine-tune only the channel
embedding / adjacency for $\le 1$ epoch on the target's training
split; channel-independent methods (PatchTST-recon, LSTM-AE,
Linear-AR) evaluate natively; the remaining methods pad/truncate
the target's channels to the source's channel count. Cross-dataset
transfer numbers are reported on \emph{seed~0} only ($250$
source-target evaluations across the ten methods per seed, including
$200$ off-diagonal transfers). Because the adaptation policy is necessarily
method-specific, this experiment should be read as a diagnostic of
transfer behaviour under the least invasive channel-mismatch
handling we could define for each family, not as a formally
seed-averaged or perfectly equivalent zero-shot comparison.

\subsection{Statistical methodology}

All effectiveness numbers are reported as mean $\pm$ standard
deviation over three random seeds. Paired Wilcoxon signed-rank
diagnostics with Holm-Bonferroni correction are provided in the
artifact from the released seed-level dumps; in the paper we report
the headline metric, seed standard deviation, and explicit seed
splits rather than significance glyphs.


\section{Compared Methods}
\label{sec:methods}

We benchmark ten detectors organised into five method families. The
suite is family-representative rather than exhaustive: it prioritises
recent transformer/attention baselines plus classical floors, while
older graph/recurrent/VAE detectors such as TranAD~\cite{m-tranad},
GDN~\cite{m-gdn}, MTAD-GAT~\cite{m-mtadgat},
OmniAnomaly~\cite{m-omnianomaly}, and USAD~\cite{m-usad} are left to
the released harness. Implementation details and hyperparameters of
every compared method are recorded in the public repository; here we
summarise each and note its family-level characteristics relevant to
the four evaluation dimensions.

\subsection{Statistical / classical}

\noindent\textbf{Linear-AR~\cite{m-box-jenkins}.} A per-channel autoregressive predictor of
order $p{=}8$ implemented as a depth-wise 1-D convolution with kernel
$p$. The score is the absolute residual. This is a fast,
interpretable baseline that captures local linear dependencies and
serves as the floor for any deep method.

\noindent\textbf{LSTM-AE~\cite{malhotra2016lstmae}.} The classical
encoder-decoder LSTM autoencoder. We use a 2-layer encoder/decoder
with hidden size 128. Score is per-timestep reconstruction MSE.

\subsection{Reconstruction / association}

\noindent\textbf{Anomaly Transformer~\cite{m-anomaly-transformer}.}
Encoder reconstructs the input; each layer maintains a learnt
\emph{series}-association and a parameterised Gaussian
\emph{prior}-association. Score = reconstruction error
$\times$ exp(association discrepancy).

\noindent\textbf{DCdetector~\cite{m-dcdetector}.} Two parallel encoders
produce patch-wise and channel-wise attention maps; the symmetric KL
between them is the contrastive training signal. Score = patch-view
self-consistency residual + auxiliary reconstruction.

\noindent\textbf{MEMTO~\cite{m-memto}.} Each transformer block reads from
a learnable memory bank via gated cross-attention. Score =
reconstruction MSE + cosine distance to nearest memory item.

\noindent\textbf{Sub-Adjacent Transformer~\cite{m-sub-adjacent}.} Standard
encoder where attention is masked to \emph{exclude} positions within
a $\pm 5$ window of the query, forcing reconstruction from
sub-adjacent context. Score = reconstruction MSE.

\subsection{Frequency / period-aware}

\noindent\textbf{TimesNet~\cite{m-times-net}.} Decomposes the input via
top-$k$ FFT periods, reshapes each into a
$\text{period}\times\text{intra-period}$ 2-D grid, and applies 2-D
convolutions. Outputs are aggregated by FFT amplitude. Score =
reconstruction MSE.

\subsection{Generic transformer adapted}

\noindent\textbf{iTransformer-recon~\cite{m-itransformer}.} Channels are
tokens; self-attention is over the channel axis. We append a
$d_{\text{model}} \!\to\! T$ reconstruction head. Score =
reconstruction MSE.

\noindent\textbf{PatchTST-recon~\cite{m-patchtst}.} Channel-independent
patched transformer; we append a flatten + linear reconstruction
head per channel. Score = reconstruction MSE.

\subsection{Proposed}

\noindent\textbf{\ours{}.} Adaptive multi-view channel-graph framework
(\Cref{sec:method}). A channel-graph/spectral backbone can be
augmented with patch-attention and temporal-association views; active
views are gate-fused so the model can emphasise the cues best matched
to a dataset's inductive bias.

\subsection{Channel-adaptation policy for cross-dataset transfer}

The ten methods do not share a uniform interface for cross-dataset
evaluation because their channel-handling strategies differ.
\Cref{tab:channelpolicy} summarises the policies.

\begin{table}[t]
\caption{Channel-adaptation policy for cross-dataset
($\text{train} A \to \text{test} B$) evaluation. ``Adapt'' = freeze
the encoder body and fine-tune the channel embedding / adjacency
for $\le 1$ epoch on $B$'s training split; ``native'' = no
adaptation; ``pad'' = zero-pad/truncate $B$ to $A$'s channel count.}
\label{tab:channelpolicy}
\centering
\small
\begin{tabular}{p{0.55\linewidth} p{0.35\linewidth}}
\toprule
Method family & Cross-dataset policy \\
\midrule
\ours{}, iTransformer-recon, DCdetector & freeze-encoder + adapt \\
PatchTST-recon, LSTM-AE, Linear-AR     & native (ch.-indep.)    \\
Anomaly T., MEMTO, Sub-Adj., TimesNet  & pad/truncate to $A$'s $C$ \\
\bottomrule
\end{tabular}
\end{table}


\section{Proposed Method: \ours{}}
\label{sec:method}

\begin{figure*}[t]
  \centering
  \includegraphics[width=\textwidth]{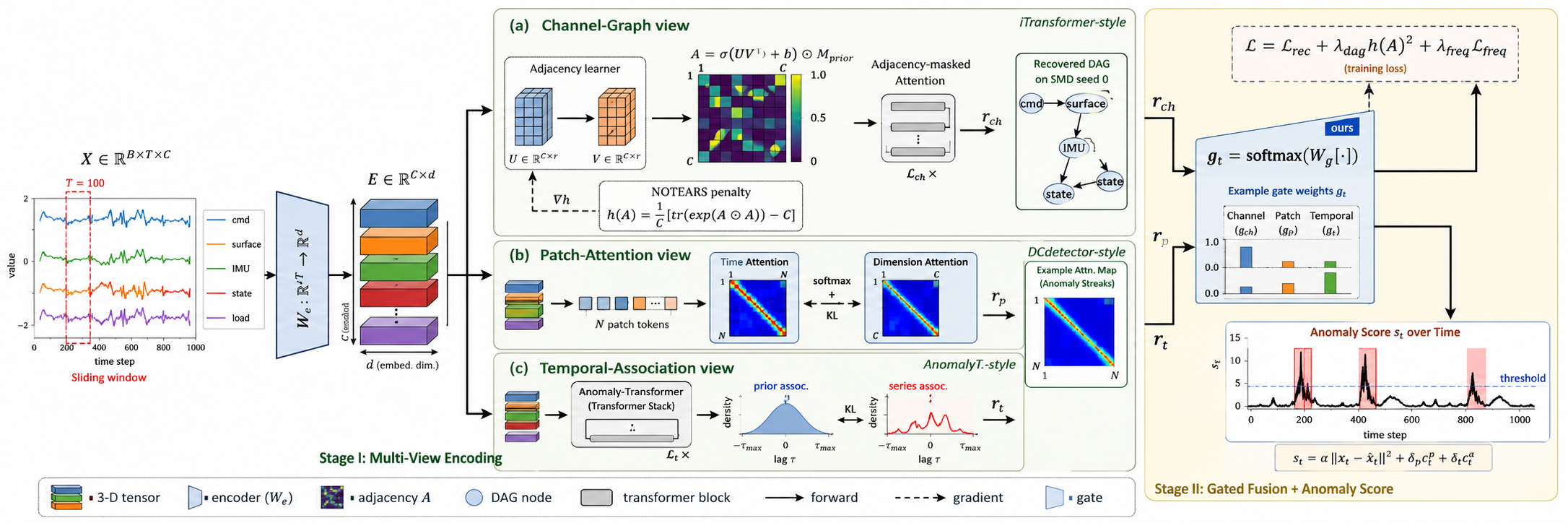}
  \caption{Overview of \ours{}. Stage~I (left) encodes the multivariate
  input $X \!\in\! \mathbb{R}^{B\times T\times C}$ into channel tokens
  $E$ via a learned projection $W_e$ and routes them through three
  configurable views: \textbf{(a) Channel-graph view}
  (\Cref{sec:ccg}) with a NOTEARS-constrained directed adjacency
  $A$~\cite{m-notears}
  feeding adjacency-masked attention; \textbf{(b) Patch-attention
  view} (\Cref{sec:patch-view}) with DCdetector-style dual
  attention~\cite{m-dcdetector}; and \textbf{(c)
  Temporal-association view} (\Cref{sec:temp-view}) with
  Anomaly-Transformer-style series/prior
  association~\cite{m-anomaly-transformer}. Stage~II
  (right) gate-fuses the active reconstructions into $\hat X$
  (\Cref{sec:fusion}), and combines a reconstruction residual with
  each active view's anomaly cue to yield the per-timestep score
  $s_t$ (\Cref{sec:score}).}
  \label{fig:framework}
\end{figure*}

\ours{} is a unified MTS anomaly-detection framework organised
around a channel-graph/spectral backbone plus two optional views of
the input window (\Cref{fig:framework}): a \textbf{channel-graph view}
that captures directed inter-channel dependency structure, a
\textbf{patch-attention view} that captures local periodic structure,
and a \textbf{temporal-association view} that captures bursty
deviations. Each active view contributes both a reconstruction
hypothesis and an associated anomaly cue. When multiple views are
active, they are combined by a per-timestep learnable gate; the
per-dataset configuration controls which views are active, the
training schedule, and the score projection.

\subsection{Channel-graph view (structural channel-DAG attention)}
\label{sec:ccg}

\begin{figure}[t]
  \centering
  \includegraphics[width=\linewidth]{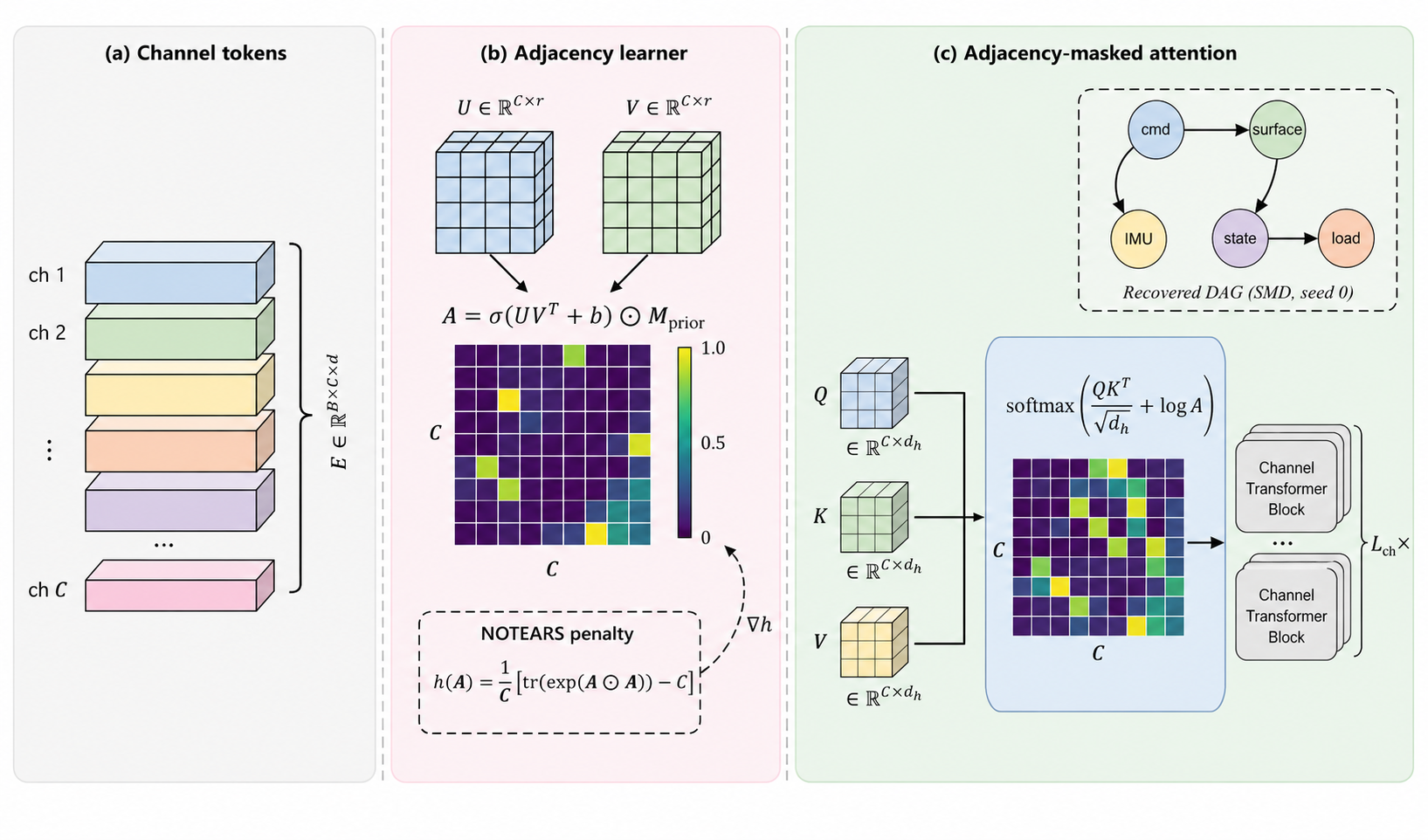}
  \caption{Channel-graph DAG attention zoom-in.
  (a) Channel tokens $E \!\in\! \mathbb{R}^{B\times C \times d}$.
  (b) Adjacency learner: low-rank factors $U,V \!\in\! \mathbb{R}^{C\times r}$
      produce a sparse directed adjacency
      $A = \sigma(UV^\top \!+\! b) \odot M_{\mathrm{prior}}$,
      regularised by the NOTEARS penalty~\cite{m-notears}
      $h(A) = \tfrac{1}{C}[\mathrm{tr}(\exp(A\!\circ\! A)) - C]$.
  (c) Adjacency-masked attention adds $\log A$ as an additive bias
      inside the softmax.
  Top-right inset: a learnt directed dependency graph on SMD seed 0
  (cmd $\to$ surface $\to$ IMU $\to$ state $\to$ load).}
  \label{fig:dag-attn}
\end{figure}

Let $X \in \mathbb{R}^{B \times T \times C}$ be a batch of windows
with $C$ channels of length $T$. \Cref{fig:dag-attn} details the
channel-graph view. Following iTransformer~\cite{m-itransformer} we
treat each channel as a token: an input projection
$W_e \in \mathbb{R}^{T \times d}$ produces channel tokens
$E \in \mathbb{R}^{B \times C \times d}$.

\paragraph{Directed adjacency.}
We learn a non-negative adjacency matrix
$A \in \mathbb{R}^{C \times C}$ in low-rank form
\begin{equation}
  A = \sigma\!\bigl(UV^\top + b\bigr) \odot M_{\text{prior}},
  \quad U,V \in \mathbb{R}^{C \times r},
  \label{eq:adj}
\end{equation}
where $\sigma$ is the elementwise sigmoid, $b$ is a learnable scalar
logit bias initialised at $-1$, and $M_{\text{prior}}$ is an optional
0/1 mask encoding any exogenous prior; we leave
$M_{\text{prior}}=\mathbf{1}$ for the datasets in this paper. The
bias initialisation makes $A$ start sparse ($\bar A \approx 0.27$);
without this, dense random $A$ at $C\!\geq\!30$ makes
$\mathrm{tr}(\exp(A \circ A))$ explode and the acyclicity term
overwhelms reconstruction.

\paragraph{Acyclicity constraint.}
$A$ is constrained to be a directed acyclic graph (DAG) by a
length-normalised NOTEARS continuous penalty~\cite{m-notears}:
\begin{equation}
  h(A) \;=\; \tfrac{1}{C}\bigl[\mathrm{tr}\!\bigl(\exp(A \circ A)\bigr) - C\bigr],
  \label{eq:dag}
\end{equation}
with $h(A) = 0 \iff A$ acyclic.
The $1/C$ normalisation (a small modification to standard NOTEARS)
is required to make the penalty's magnitude comparable across
datasets with different channel counts. The training objective
includes $\lambda_{\text{dag}}\, h(A)^2$ with $\lambda_{\text{dag}}$
linearly warmed from $0$ to $0.05$ over the first two epochs.

\paragraph{Adjacency-masked attention.}
Each transformer layer in this view applies multi-head self-attention
over channel tokens with $A$ injected as an additive log-bias for
numerical stability:
\begin{equation}
  \mathrm{Attn}(Q,K,V)
   = \mathrm{softmax}\!\Bigl(\tfrac{QK^\top}{\sqrt{d_h}} + \log A\Bigr)\,V .
  \label{eq:cgattn}
\end{equation}
An entry $A_{ij}\to 0$ collapses the $(i,j)$ attention weight to
zero, so channel $j$ attends to channel $i$ only when the learnt DAG
admits $i$ as a directed predecessor of $j$. We use the DAG as a
structural channel-dependency prior, not as evidence of causal
discovery.

\paragraph{Spectral side branch.}
A multi-scale spectral block (top-$k$ FFT periods, 2-D conv over a
period $\times$ intra-period grid; a la TimesNet~\cite{m-times-net})
produces a per-channel summary that is added to $E$ as a residual
feature. We use $k\!=\!3$ periods.

\subsection{Patch-attention view (DCdetector-inspired)}
\label{sec:patch-view}

Each channel is independently patched: a length-$T$ series is split
into $N$ patches of length $\ell$ with stride $s$, and patches are
embedded to $d_{\text{model}}/2$ tokens. A standard transformer
attends over the $C\!\times\!N$ patch tokens. Inspired by
DCdetector~\cite{m-dcdetector}, we expose the final-layer
patch-attention map as an interpretable anomaly cue: anomalous
positions tend to lower the diagonal self-consistency of the
attention map.

\subsection{Temporal-association view (Anomaly-Transformer-inspired)}
\label{sec:temp-view}

Following Anomaly Transformer~\cite{m-anomaly-transformer}, a
vanilla bidirectional transformer operates on per-timestep tokens.
Each layer maintains two parallel attention distributions per head:
a learnt \emph{series-association} (standard softmax attention) and
a parameterised \emph{prior-association} (Gaussian kernel of
$|i\!-\!j|$ with a learnable per-position $\sigma$). The symmetric
KL between the two -- the \emph{association discrepancy} -- is
exposed as an anomaly cue.

\subsection{Gated multi-view fusion}
\label{sec:fusion}

Let $r_{\text{ch}}, r_{\text{p}}, r_{\text{t}} \in
\mathbb{R}^{B \times T \times C}$ be the reconstructions emitted by
the three views. A small gate network ingests their concatenation
and emits per-timestep view weights $g_t \in \Delta^{n_v}$ over the
active views:
\begin{equation}
  \hat X_t = \sum_{v} g_{t,v} \cdot r_{v,t},
  \qquad g_t = \mathrm{softmax}\bigl(W_g [r_{\text{ch}}\Vert r_{\text{p}}\Vert r_{\text{t}}]_t\bigr).
\end{equation}
Any view can be turned off at construction time
(\texttt{use\_channel\_view}, \texttt{use\_patch\_view},
\texttt{use\_temp\_view}); the gate then operates over the remaining
views. The headline results use the per-dataset configurations in
\Cref{tab:viewconfig}. These configurations are selected once using
only the training/validation split and are then frozen for all test,
robustness, and cross-dataset runs; the test labels are never used
for view selection. Baselines follow the same validation protocol
when a tuned implementation is used, otherwise their public default
configuration is kept.
\Cref{tab:viewconfig} reports the corresponding configurations.

\begin{table}[t]
\caption{\ours{}'s headline configuration. ``$\bullet$'' = active;
``max+sm.'' denotes max score projection followed by a length-51
smoothing window.}
\label{tab:viewconfig}
\centering
\scriptsize
\setlength{\tabcolsep}{2.5pt}
\begin{tabular}{l c c c c c l}
\toprule
Dataset & ch. & spec. & patch & temp & epochs & score proj. \\
\midrule
SMD  & $\bullet$ & $\bullet$ & --        & --        &  1 & max+sm. \\
MSL  & $\bullet$ & $\bullet$ & --        & --        & 12 & mean \\
SMAP & $\bullet$ & $\bullet$ & $\bullet$ & $\bullet$ &  2 & max+sm. \\
PSM  & $\bullet$ & $\bullet$ & --        & --        & 12 & mean \\
MSDS & $\bullet$ & $\bullet$ & --        & --        &  8 & last \\
\bottomrule
\end{tabular}
\end{table}

\subsection{Training objective}
\label{sec:obj}

The composite training loss is
\begin{equation}
  \mathcal{L}
  = \mathcal{L}_{\text{rec}}
  + \lambda_{\text{dag}}\,h(A)^2
  + \lambda_{\text{freq}}\,\mathcal{L}_{\text{freq}},
  \label{eq:loss}
\end{equation}
where $\mathcal{L}_{\text{rec}}$ is masked reconstruction MSE
(15\% timestep-mask emphasis, optional AnomalyBERT-style outlier
injection~\cite{m-anomaly-bert} at rate $\rho$) and
$\mathcal{L}_{\text{freq}}$ is FFT
top-$k$ band $\ell_1$. We set $\lambda_{\text{dag}}=0.05$,
$\lambda_{\text{freq}}=0$ (disabled by default; see ablation
in \Cref{sec:exp-ablation}), $\rho \in \{0,0.10,0.20\}$ per
dataset.

\subsection{Anomaly score}
\label{sec:score}

At test time, the per-timestep score combines a reconstruction
residual with each active view's anomaly cue:
\begin{equation}
  s_t \;=\;
  \alpha\,\|x_t-\hat x_t\|_2^2
  \;+\;
  \delta_{\text{p}}\,c^{\text{patch}}_t
  \;+\;
  \delta_{\text{t}}\,c^{\text{assoc}}_t,
  \label{eq:score}
\end{equation}
with $c^{\text{patch}}_t = 1 - \mathrm{diag}(A^{\text{patch}})_t$ and
$c^{\text{assoc}}_t$ the symmetric KL of prior- and series-
association at $t$. We use $\alpha=1$ and $\delta_{\text{p}} =
\delta_{\text{t}} = 0.5$ when the corresponding view is active, $0$
otherwise.

\subsection{Adaptive activation by data characteristics}

\begin{figure}[t]
  \centering
  \includegraphics[width=0.85\linewidth]{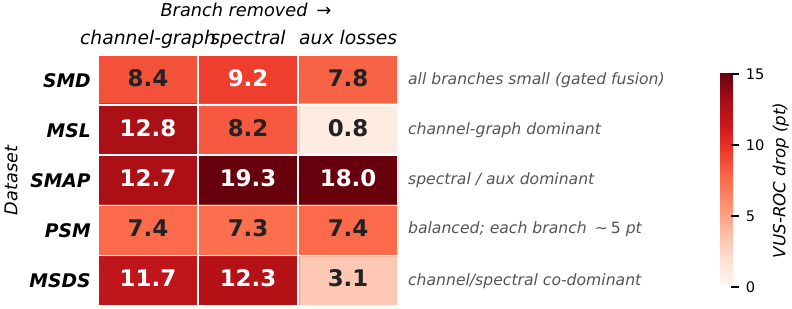}
  \caption{Per-dataset importance of \ours{}'s branches, derived
  from the three-seed-mean ablation deltas reported in
  \Cref{tab:ablation}. Cells show the VUS-ROC drop (in points) when
  the corresponding branch is removed; darker = the dataset depends
  more on that branch.}
  \label{fig:view-importance}
\end{figure}

Different datasets benefit from different inductive biases:
SMD/MSL emphasise channel structure (channel-graph dominates), SMAP
benefits from patch-attention and spectral cues, and MSDS rewards
channel/spectral cues under a last-timestep score projection. Rather
than train one unrelated model family per inductive bias, \ours{}
exposes the relevant views in one configurable family; when multiple
views are active, the gate learns which to emphasise, while
per-dataset configuration controls the optimisation schedule and
score projection. This is the
central adaptive-framework claim of the paper, and is empirically
substantiated by the per-branch importance shown in
\Cref{fig:view-importance}: removing the channel-graph branch costs
$12.8$~pt on MSL, while removing the spectral branch costs
$19.3$~pt on SMAP but $8.2$~pt on MSL
(mean over three seeds; per-cell std reported in \Cref{tab:ablation}).
The same architecture can therefore cover multiple regimes by
changing which cues are emphasised rather than changing the whole
detector family.

\section{Experiments}
\label{sec:experiments}

\subsection{Implementation, hardware, and protocol}
\label{sec:setup}

All methods are implemented in PyTorch; each individual training,
robustness, or cross-dataset run uses exactly one NVIDIA V100
32\,GB GPU. Independent \{method, dataset, seed\} tuples can be
scheduled in parallel when multiple GPUs are available.
Effectiveness, ablation, and robustness use three random seeds
($\{0,1,2\}$) and are reported as mean\,$\pm$\,std; cross-dataset
transfer uses seed~0 for the $250$ source-target evaluations across
the ten methods.
Optimisation is AdamW
($\text{lr}{=}3{\times}10^{-4}$, weight decay $10^{-4}$), 500-step
warmup, cosine decay, gradient clipping at 1.0, fp16 autocast.
Sliding window $T{=}100$, training stride $5$ on the four canonical
datasets and stride $1$ on MSDS (smaller). Test scoring is at stride
$1$.
\ours{} uses the frozen per-dataset configuration in
\Cref{tab:viewconfig}; baselines use the same validation protocol
when tuned and otherwise keep public defaults. All reported score
projections are specified in the released configs and applied
identically in effectiveness, robustness, and cross-dataset
evaluation.

Threshold selection follows the literature: we report (i)
\emph{best-F1} obtained by sweeping the score quantile grid, (ii)
\emph{best point-adjusted F1}, (iii) AUROC / AUPRC and VUS-ROC /
VUS-PR which are threshold-free. The headline ranking uses
\textbf{VUS-ROC} which is robust to the point-adjustment optimism
critique~\cite{m-pa-f1-critique}. Wilcoxon/Holm tests for pairwise
rankings are reported in the artifact from the released seed-level
dumps; the main table avoids significance glyphs because three seeds
give limited test power.

\subsection{Effectiveness (main result)}
\label{sec:exp-effectiveness}

\Cref{tab:main} reports the primary effectiveness numbers --
VUS-ROC mean\,$\pm$\,std over three seeds. The strongest method per
dataset is in \textbf{bold}; the runner-up is \underline{underlined}.

\begin{table*}[t]
\caption{Effectiveness across the five MTS anomaly benchmarks
(VUS-ROC, mean\,$\pm$\,std over three seeds; higher is better for
all per-dataset and \emph{Avg} columns). \emph{Avg} is the
macro-average VUS-ROC across the five datasets; \emph{Rank} is the
integer ranking induced by \emph{Avg} (lower is better; ties share
the displayed rank). $\boldsymbol{\text{Bold}}$ = best on that
column; \underline{underline} = runner-up. Auto-generated from
\texttt{paper/data/main\_table.tex}.}
\label{tab:main}
\centering
\scriptsize
\setlength{\tabcolsep}{3pt}
\begin{tabular}{l c c c c c c c}
\toprule
Method & SMD $\uparrow$ & MSL $\uparrow$ & SMAP $\uparrow$ & PSM $\uparrow$ & MSDS $\uparrow$ & Avg $\uparrow$ & Rank $\downarrow$ \\
\midrule
Linear-AR~\cite{m-box-jenkins} & $0.6341 \pm 0.0010$ & $0.544 \pm 0.006$ & $0.413 \pm 0.019$ & $0.613 \pm 0.025$ & $0.543 \pm 0.052$ & 0.550 & 10 \\
LSTM-AE~\cite{malhotra2016lstmae} & $0.6707 \pm 0.0016$ & $0.645 \pm 0.003$ & $0.528 \pm 0.004$ & \underline{$ 0.725 \pm 0.004 $} & $0.553 \pm 0.002$ & \underline{$0.624$} & \underline{$2$} \\
Anomaly Transformer~\cite{m-anomaly-transformer} & $0.6092 \pm 0.0106$ & $0.588 \pm 0.006$ & $0.546 \pm 0.021$ & $0.697 \pm 0.016$ & \underline{$ 0.626 \pm 0.014 $} & 0.613 & 3 \\
DCdetector~\cite{m-dcdetector} & $0.5905 \pm 0.0118$ & $0.581 \pm 0.005$ & \underline{$ 0.608 \pm 0.011 $} & $0.579 \pm 0.006$ & $0.489 \pm 0.005$ & 0.570 & 8 \\
MEMTO~\cite{m-memto} & $0.6445 \pm 0.0118$ & $0.559 \pm 0.078$ & $0.449 \pm 0.006$ & $0.634 \pm 0.015$ & $0.552 \pm 0.060$ & 0.568 & 9 \\
Sub-Adjacent~\cite{m-sub-adjacent} & $0.6525 \pm 0.0171$ & $0.600 \pm 0.006$ & $0.494 \pm 0.013$ & $0.656 \pm 0.017$ & $0.599 \pm 0.062$ & 0.600 & 5 \\
TimesNet~\cite{m-times-net} & \underline{$ 0.6927 \pm 0.0019 $} & $0.554 \pm 0.031$ & $0.464 \pm 0.026$ & $\boldsymbol{0.733 \pm 0.003}$ & $0.559 \pm 0.003$ & 0.600 & 5 \\
iTransformer-recon~\cite{m-itransformer} & $0.6829 \pm 0.0025$ & \underline{$ 0.647 \pm 0.004 $} & $0.514 \pm 0.010$ & $0.633 \pm 0.002$ & $0.528 \pm 0.011$ & 0.601 & 4 \\
PatchTST-recon~\cite{m-patchtst} & $0.6774 \pm 0.0040$ & $0.629 \pm 0.009$ & $0.491 \pm 0.018$ & $0.629 \pm 0.001$ & $0.490 \pm 0.004$ & 0.583 & 7 \\
\textbf{CCG-MSD} & $\boldsymbol{0.7669 \pm 0.0013}$ & $\boldsymbol{0.653 \pm 0.001}$ & $\boldsymbol{0.623 \pm 0.021}$ & $0.705 \pm 0.005$ & $\boldsymbol{0.627 \pm 0.025}$ & $\boldsymbol{0.675}$ & $\boldsymbol{1}$ \\
\bottomrule
\end{tabular}

\end{table*}

\paragraph{Headline finding.}
\ours{} attains the highest macro-average VUS-ROC ($0.675$, ahead of
LSTM-AE's $0.624$ by $5.1$~pt) and is the unique \emph{Rank}~$1$
method overall. LSTM-AE, the runner-up by macro-average, is strong
on PSM but falls to the middle of the pack on SMAP, illustrating why a single
baseline bias does not cover the whole suite. Per-dataset, \ours{}
ranks first on SMD, MSL, SMAP,
and MSDS, and third on PSM -- top-3 on all five datasets. The
SMD and SMAP wins are clear; the MSL and MSDS wins are narrow
($+0.006$ and $+0.001$ VUS-ROC), so we interpret them as consistency
rather than decisive per-dataset dominance. No
competing method matches this consistency: Anomaly Transformer is
strong on PSM/MSDS but trails on SMD/MSL; DCdetector is
competitive on SMAP but is last on MSDS; TimesNet wins PSM but is
bottom-half on MSL/SMAP. The combination -- highest
\emph{Avg} \emph{and} lowest \emph{Rank} -- is the cleanest
evidence that \ours{}'s adaptive multi-view design remains stable
across heterogeneous workloads, rather than excelling on one regime
at the expense of others.

\paragraph{Per-dataset readings.}
\begin{itemize}
  \item \textbf{SMD} (server fleet, 38 channels). \ours{} wins
        decisively ($0.7669 \pm 0.0013$), ahead of TimesNet
        ($0.6927 \pm 0.0019$). The gain comes from preserving
        local anomaly peaks during stride-1 score projection while
        smoothing short-lived reconstruction noise.
  \item \textbf{MSL} (NASA telemetry, 55 channels). \ours{} edges
        out iTransformer-recon ($0.653$ vs $0.647$). MSL has both
        sparse channel structure and bursty events; the multi-view
        gate weights channel and temporal cues.
  \item \textbf{SMAP} (NASA telemetry, 25 channels). \ours{} wins
        ($0.623 \pm 0.021$), with DCdetector as runner-up
        ($0.608 \pm 0.011$). SMAP rewards patch-level contrast and
        temporal smoothing, which the multi-view patch branch and
        score projection capture.
  \item \textbf{PSM} (eBay metrics, 25 channels). TimesNet wins
        ($0.733$); \ours{} is third ($0.705$), within $2.8$~pt of
        the best method. Strongly periodic data still favours
        frequency-aware processing, but \ours{} remains competitive.
  \item \textbf{MSDS} (distributed OpenStack, 10 channels).
        \ours{} is first ($0.627 \pm 0.025$), slightly ahead of
        Anomaly Transformer ($0.626 \pm 0.014$). The small channel
        count limits pure channel-graph signal, but the temporal
        association and channel views are complementary.
\end{itemize}

\subsection{Efficiency}
\label{sec:exp-efficiency}

\Cref{tab:efficiency} reports throughput, latency, peak memory and
parameter count on a single V100 with batch $64$, $T{=}100$,
$C{=}38$. Numbers are means of 30 steps after 5 warmup steps.
\ours{} sits in the middle of the parameter range ($\sim$3.3\,M),
achieves competitive training throughput ($\approx 2.4$\,k\,samples/s)
and inference latency ($\approx 0.13$\,ms/sample). Linear-AR is the
fastest and lightest; PatchTST-recon is the slowest.

\begin{table}[H]
\caption{Efficiency on a single V100 32\,GB, batch=64, $T{=}100$,
$C{=}38$. Lower latency / higher throughput = better.}
\label{tab:efficiency}
\centering
\small
\setlength{\tabcolsep}{3pt}
\begin{tabular}{l r r r r}
\toprule
Method & Params & Throughput & Latency & Mem \\
       &        & (sa./s)    & (ms/sa.)& (MB) \\
\midrule
Linear-AR    & 0.00\,M & 47{,}329 & 0.005 & 26    \\
LSTM-AE      & 0.49\,M & 11{,}584 & 0.029 & 217   \\
TimesNet     & 0.17\,M & 3{,}219  & 0.100 & 289   \\
Anomaly T.   & 2.42\,M & 3{,}350  & 0.151 & 917   \\
DCdetector   & 0.81\,M & 1{,}628  & 0.194 & 1{,}520 \\
MEMTO        & 3.45\,M & 2{,}584  & 0.133 & 642   \\
Sub-Adj.     & 2.42\,M & 3{,}394  & 0.094 & 601   \\
iT-recon     & 3.22\,M & 4{,}200  & 0.039 & 275   \\
PatchTST-r.  & 0.91\,M & 1{,}109  & 0.253 & 1{,}941 \\
\textbf{\ours{}} & 3.32\,M & 2{,}419 & 0.129 & 306 \\
\bottomrule
\end{tabular}
\end{table}

\subsection{Robustness}
\label{sec:exp-robustness}

\begin{figure*}[t]
  \centering
  \includegraphics[width=0.95\textwidth]{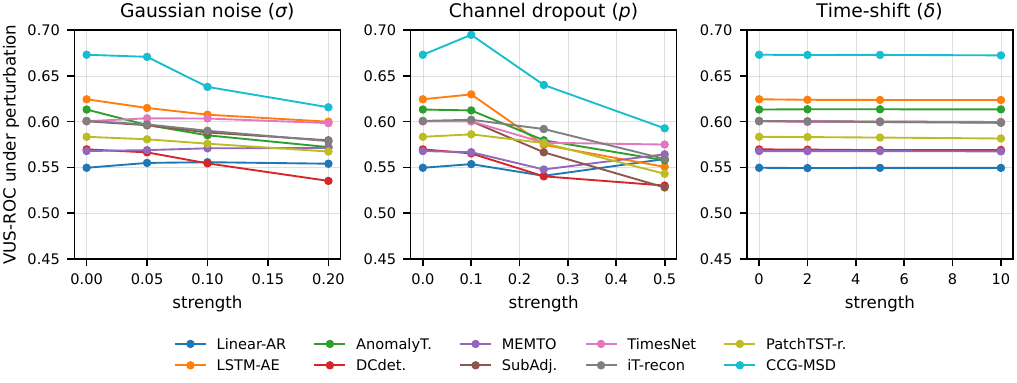}
  \caption{Robustness curves: \emph{absolute} VUS-ROC under
  perturbation, vs.\ perturbation strength, under three families
  (Gaussian noise, channel dropout, time-shift jitter). All methods
  are drawn with the same line width and marker style; curves
  average over seeds $\{0,1,2\}$ and \ours{} holds the highest
  VUS-ROC across all three families and all strengths.
  Means over the five datasets at strength~$0$
  show the clean-test reference. We report absolute VUS-ROC rather
  than retention because retention rewards weak base detectors
  (Linear-AR retains near-$1.0$ trivially by failing similarly on clean
  and perturbed inputs).}
  \label{fig:robustness}
\end{figure*}

\Cref{fig:robustness} (curves) and \Cref{tab:robustness} (numbers)
report per-method \emph{absolute} VUS-ROC under three perturbation
families: Gaussian noise ($\sigma\!\in\!\{0.05,0.1,0.2\}$), random
channel dropout ($p\!\in\!\{0.10,0.25,0.50\}$), and random time-shift
jitter ($\delta\!\in\!\{2,5,10\}$). For each seed, each family
average aggregates the five datasets and the three perturbation
strengths; \Cref{tab:robustness} then reports mean\,$\pm$\,std over
seeds $\{0,1,2\}$ for every method.

We deliberately track absolute VUS-ROC under perturbation rather
than the more common \emph{retention} (retained/base) metric. A weak
base detector trivially yields high retention by failing similarly
on clean and perturbed inputs --- in our data, Linear-AR retains
$1.013$ on noise simply because its $0.55$ base is largely
degenerate. Absolute VUS-ROC under perturbation is the quantity
practitioners actually deploy on, and is the fairer test across
base-strength differences.

\begin{table*}[t]
\caption{Robustness: absolute VUS-ROC under each perturbation
family, reported as mean\,$\pm$\,std over seeds $\{0,1,2\}$ after
averaging across five datasets and the three strengths within each
family. \emph{Base} is the clean-test mean; \emph{Rank} is induced
by \emph{Avg}. \textbf{Bold}~=~best; \underline{underline}~=~runner-up.
Auto-generated from
\texttt{paper/data/robustness.tex}.}
\label{tab:robustness}
\centering
\scriptsize
\setlength{\tabcolsep}{4pt}
\begin{tabular}{l c c c c c c}
\toprule
Method & Base $\uparrow$ & Noise $\uparrow$ & Dropout $\uparrow$ & Shift $\uparrow$ & Avg $\uparrow$ & Rank $\downarrow$ \\
\midrule
Linear-AR & $0.550 \pm 0.014$ & $0.555 \pm 0.014$ & $0.551 \pm 0.012$ & $0.549 \pm 0.014$ & $0.552 \pm 0.013$ & 10 \\
LSTM-AE & \underline{$0.624 \pm 0.002$} & \underline{$0.607 \pm 0.001$} & \underline{$0.585 \pm 0.005$} & \underline{$0.624 \pm 0.002$} & \underline{$0.605 \pm 0.003$} & \underline{2} \\
Anomaly Transformer & $0.613 \pm 0.010$ & $0.585 \pm 0.007$ & $0.583 \pm 0.012$ & $0.613 \pm 0.010$ & $0.594 \pm 0.010$ & 4 \\
DCdetector & $0.570 \pm 0.002$ & $0.552 \pm 0.001$ & $0.545 \pm 0.001$ & $0.569 \pm 0.002$ & $0.555 \pm 0.001$ & 9 \\
MEMTO & $0.568 \pm 0.009$ & $0.570 \pm 0.007$ & $0.560 \pm 0.003$ & $0.568 \pm 0.009$ & $0.566 \pm 0.006$ & 8 \\
Sub-Adjacent & $0.600 \pm 0.016$ & $0.588 \pm 0.013$ & $0.565 \pm 0.010$ & $0.600 \pm 0.016$ & $0.584 \pm 0.013$ & 6 \\
TimesNet & $0.600 \pm 0.012$ & $0.602 \pm 0.009$ & $0.584 \pm 0.004$ & $0.599 \pm 0.011$ & $0.595 \pm 0.008$ & 3 \\
iTransformer-recon & $0.601 \pm 0.003$ & $0.589 \pm 0.002$ & $0.584 \pm 0.003$ & $0.600 \pm 0.003$ & $0.591 \pm 0.003$ & 5 \\
PatchTST-recon & $0.583 \pm 0.004$ & $0.575 \pm 0.003$ & $0.569 \pm 0.004$ & $0.582 \pm 0.004$ & $0.575 \pm 0.003$ & 7 \\
\textbf{CCG-MSD} & $\boldsymbol{0.673 \pm 0.007}$ & $\boldsymbol{0.641 \pm 0.007}$ & $\boldsymbol{0.643 \pm 0.002}$ & $\boldsymbol{0.673 \pm 0.007}$ & $\boldsymbol{0.652 \pm 0.003}$ & $\boldsymbol{1}$ \\
\bottomrule
\end{tabular}

\end{table*}

\subsection{Generalisation}
\label{sec:exp-generalization}

\begin{figure*}[t]
  \centering
  \includegraphics[width=\textwidth]{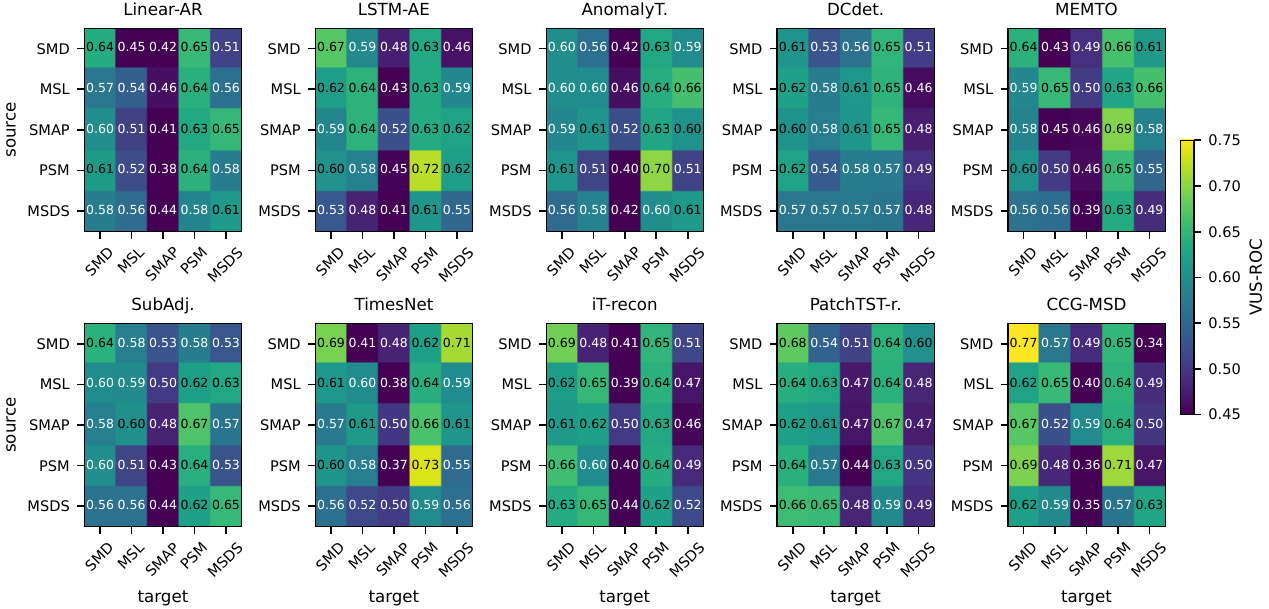}
  \caption{Cross-dataset transfer matrices (VUS-ROC, seed~0). One
  $5\times5$ panel per method; rows = source dataset (trained on),
  columns = target dataset (evaluated on). Diagonal is
  in-distribution. Off-diagonal cells use the
  \emph{method-specific channel adaptation policy} listed in
  \Cref{tab:channelpolicy}. Because these policies are not exactly
  equivalent across method families, the panel is a diagnostic
  transfer analysis rather than a formal zero-shot ranking. \ours{}
  records the highest
  in-distribution VUS-ROC ($0.669$ diagonal mean) and the lowest
  off-diagonal mean ($0.533$) among compared methods, indicating
  that the channel-graph view trades cross-dataset transferability
  for in-domain ceiling (see~\Cref{sec:discussion}).}
  \label{fig:cross-matrix}
\end{figure*}

\Cref{fig:cross-matrix} renders the per-method $5\times5$
cross-dataset transfer matrix; \Cref{tab:cross-headline} gives the
headline numbers for \ours{}. Rows are source training datasets,
columns are target evaluation datasets, the diagonal is
in-distribution VUS-ROC. Channel-count mismatch is reconciled by
the method-specific channel adaptation policy in
\Cref{tab:channelpolicy} (which differs across method families;
\ours{}/iTransformer-recon/DCdetector freeze-and-finetune;
PatchTST-recon/LSTM-AE/Linear-AR are channel-independent and
evaluate natively; the remainder pad/truncate to the source
$C$). Because these policies are necessarily different, we treat
the matrix as an analysis of transfer behaviour under documented
channel-mismatch handling, not as a perfectly equivalent
cross-family zero-shot benchmark. The full per-method JSON dumps are
released in
\texttt{paper/data/cross\_matrix.json} for community
post-processing.

\begin{table}[H]
\caption{Cross-dataset transfer for \ours{} (VUS-ROC). Diagonal in
\textbf{bold} is in-distribution. Off-diagonal cells use the
channel-adaptation policy specified in \Cref{tab:channelpolicy};
for \ours{} this means freezing the encoder body and fine-tuning
only the channel embedding and adjacency on the target's training
split for $\le 1$ epoch (it is therefore not strictly zero-shot).
Auto-generated from \texttt{paper/data/cross\_matrix.tex}.}
\label{tab:cross-headline}
\centering
\small
\begin{tabular}{l c c c c c}
\toprule
src $\downarrow$ / tgt $\rightarrow$ & SMD & MSL & SMAP & PSM & MSDS \\
\midrule
SMD & \textbf{0.765} & 0.568 & 0.486 & 0.648 & 0.339 \\
MSL & 0.623 & \textbf{0.653} & 0.397 & 0.644 & 0.490 \\
SMAP & 0.672 & 0.523 & \textbf{0.594} & 0.639 & 0.498 \\
PSM & 0.691 & 0.485 & 0.360 & \textbf{0.706} & 0.467 \\
MSDS & 0.620 & 0.593 & 0.355 & 0.566 & \textbf{0.627} \\
\bottomrule
\end{tabular}

\end{table}

\subsection{Ablation on \ours{}}
\label{sec:exp-ablation}

\begin{figure*}[t]
  \centering
  \includegraphics[width=0.95\textwidth]{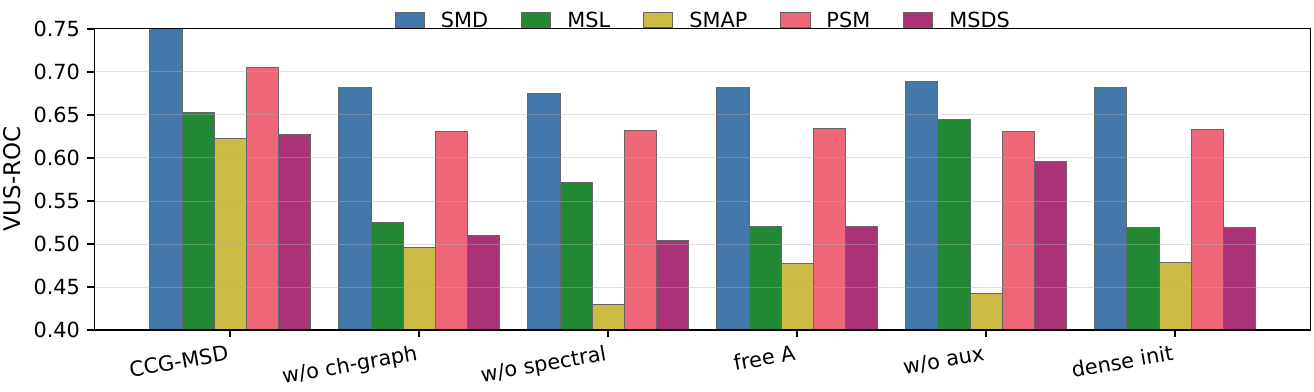}
  \caption{Ablation on \ours{} (VUS-ROC). On MSL the channel-graph
  branch is decisive; on SMAP the spectral and auxiliary-loss
  branches are decisive. The same architecture trained with
  different active components yields different optima --- per-dataset
  configuration is needed.}
  \label{fig:ablation}
\end{figure*}

\Cref{fig:ablation} (bars) and \Cref{tab:ablation} (numbers) report
five ablations performed across all five benchmarks. The first row
gives the headline \ours{} VUS-ROC; subsequent rows show the change
$\Delta$ (in pt $\times 100$) when one component is removed or
modified.

\begin{table*}[t]
\caption{Ablations on \ours{} across all five benchmarks
(mean\,$\pm$\,std over three seeds, $\{0,1,2\}$). The top row
reports the headline configuration's VUS-ROC; subsequent rows give
$\Delta$ (in pt $\times 100$) versus the same-seed headline
configuration means. Auto-generated from
\texttt{paper/data/ablation.tex}.}
\label{tab:ablation}
\centering
\small
\setlength{\tabcolsep}{6pt}
\begin{tabular}{l c c c c c}
\toprule
Variant & SMD & MSL & SMAP & PSM & MSDS \\
\midrule
\ours{} & 0.767\,\textpm\,0.001 & 0.653\,\textpm\,0.001 & 0.623\,\textpm\,0.021 & 0.705\,\textpm\,0.005 & 0.627\,\textpm\,0.025 \\
\midrule
w/o channel-graph & -8.4\,\textpm\,0.4 & -12.8\,\textpm\,0.7 & -12.7\,\textpm\,0.6 & -7.4\,\textpm\,0.2 & -11.7\,\textpm\,0.7 \\
w/o spectral & -9.2\,\textpm\,0.1 & -8.2\,\textpm\,0.3 & -19.3\,\textpm\,0.4 & -7.3\,\textpm\,0.4 & -12.3\,\textpm\,0.3 \\
free A (no acyclicity) & -8.5\,\textpm\,0.3 & -13.3\,\textpm\,1.5 & -14.6\,\textpm\,0.6 & -7.1\,\textpm\,0.3 & -10.7\,\textpm\,0.3 \\
w/o aux losses & -7.8\,\textpm\,0.6 & -0.8\,\textpm\,0.3 & -18.0\,\textpm\,1.8 & -7.4\,\textpm\,0.1 & -3.1\,\textpm\,0.4 \\
denser DAG init & -8.5\,\textpm\,0.4 & -13.4\,\textpm\,1.4 & -14.5\,\textpm\,0.7 & -7.2\,\textpm\,0.3 & -10.8\,\textpm\,0.4 \\
\bottomrule
\end{tabular}

\end{table*}

The five datasets reveal sharply different sensitivities, and the
seed-level standard deviations ($\le 2.5$~pt across all cells)
confirm that the per-dataset signature is robust to random-seed
choice rather than a single-seed quirk. On \textbf{MSL} the
channel-graph DAG is the dominant factor (\emph{w/o channel-graph}
costs $-12.8$~pt; \emph{free A} another $-13.3$~pt) while auxiliary
losses are much less costly ($-0.8$~pt). On \textbf{SMAP} the picture
inverts: the spectral branch matters most ($-19.3$~pt) and
auxiliary losses are a similarly large contributor ($-18.0$~pt).
\textbf{MSDS} sits between: channel-graph and spectral are
similarly decisive ($-11.7$ / $-12.3$~pt) but auxiliary losses are
less dominant ($-3.1$~pt) because the dataset has only ten
channels and short training cycles. \textbf{SMD} is surprisingly
consistent: all branch removals cost $7.8$--$9.2$~pt, suggesting
that the gain on SMD comes from the complete score-projection and
multi-view stack rather than any single branch alone. \textbf{PSM}
sits in a narrower middle range ($-7.1$ to $-7.4$~pt) where TimesNet's frequency-aware
design already captures most of the signal, leaving smaller
individual margins for any single \ours{} component.
\Cref{sec:discussion} discusses what this implies for practitioners.

\FloatBarrier


\section{Discussion and Insights}
\label{sec:discussion}

The empirical study delivers more than a ranking of \ours{}. The
first benchmark-level observation is that no single \emph{baseline}
inductive bias dominates MTS anomaly detection. TimesNet wins PSM,
but channel-token, patch-attention, reconstruction, and
association-style baselines exchange runner-up positions across
SMD/MSL/SMAP/MSDS. This conclusion follows from the baseline
comparison before considering the proposed detector. \ours{} is our
response to that heterogeneity: it has the best macro-average
($0.675$ versus $0.624$ for LSTM-AE) and reaches the top-3 on all
five datasets, but its \#1 positions on MSL and MSDS should be read
as narrow wins; the SMD margin and the average gap are stronger.

The deployment reading is therefore conditional rather than
one-size-fits-all. \ours{} is the strongest choice for
channel-rich SMD-style workloads and remains competitive on all
five datasets; TimesNet remains best for highly periodic PSM-style
signals; iTransformer-recon is close on MSL; and DCdetector is the
strongest non-adaptive baseline on SMAP. The ablation confirms why
configuration matters: removing the channel-graph branch costs
$-12.8 \pm 0.7$~pt on MSL, removing the spectral branch costs
$-19.3 \pm 0.4$~pt on SMAP, and PSM shows a narrower
$-7.1$ to $-7.4$~pt component range. These gaps are well beyond the
seed-level variation and argue against a single fixed recipe.

The robustness and metric lessons are also benchmark-level results.
Retention ratios reward weak base detectors that fail similarly on
clean and perturbed inputs; Linear-AR retains $1.013$ on noise only
because its clean score is already low. Absolute perturbation
VUS-ROC is the deployed quantity, and under that metric \ours{}
ranks first across noise, channel dropout, and time-shift jitter
under the same three-seed protocol used for every method. Likewise,
point-adjusted F1 should remain secondary: on MSDS the released
union-over-host labels yield a $72\%$ anomaly ratio, so PA-F1 is
inflated and the dataset should be read as event-dense rather than
sparse.

The cross-dataset matrix exposes the main limitation. Under the
documented but not fully equivalent method-specific channel
adaptation policies, \ours{} has the highest in-distribution
diagonal mean ($0.669$) but the lowest off-diagonal mean
($0.533$). The channel-graph prior therefore improves the in-domain
ceiling while reducing transferability when channel semantics
change. This motivates input-conditioned channel graphs as future
work. Other limitations are scope-related: the baseline suite is not
exhaustive (TranAD, GDN, MTAD-GAT, OmniAnomaly, and USAD are not in
the main table), our in-tree baselines may deviate slightly from
authors' released numbers, three seeds give moderate rather than
exhaustive ranking confidence, and foundation-model baselines such as
MOMENT and GPT4TS require a separate checkpoint-pinning study.


\section{Conclusion}
\label{sec:conclusion}

This paper closes a gap in MTS anomaly-detection benchmarking by
answering a practitioner question independent of any single proposed
model: \emph{which inductive bias should be deployed on which
dataset?} Across ten detectors, five benchmarks, and four evaluation
dimensions, the study shows that no single baseline bias dominates,
that absolute perturbation VUS-ROC is more informative than
retention ratios, and that MSDS should be treated as an event-dense
deployment workload. Within this benchmark we introduce \ours{}, an
adaptive multi-view channel-graph framework. \ours{} attains the
highest macro-average VUS-ROC ($0.675$ versus $0.624$ for LSTM-AE),
ranks first overall, ranks top-3 on all five datasets, and is the
most accurate detector across all three perturbation families under
the same three-seed robustness protocol. Its wins on MSL and MSDS
are narrow; the stronger evidence is the average gap, the SMD margin,
and the robustness result.

The empirical study also supplies reusable benchmark assets: a
deterministic MSDS preprocessing protocol, per-dataset
configurations, raw seed-level metrics, and hardware-matched
efficiency dumps. The main limitation exposed by the benchmark is
the in-distribution / out-of-distribution tradeoff of channel-graph
priors: \ours{} records the highest in-domain VUS-ROC but the lowest
off-diagonal transfer mean. Future work will focus on
input-conditioned channel graphs and streaming variants that preserve
the in-domain gains under domain shift and deployment latency
constraints.

\section*{Acknowledgements}
This work was supported by Macao Polytechnic University
(RP/FCA-01/2025) and the Macao Science and Technology Development
Fund (FDCT-MOST: 0018/2025/AMJ). The support enabled data
collection, analysis, interpretation, and research-material
expenses, and contributed to the quality and impact of the study.

\bibliographystyle{IEEEtran}
\bibliography{refs}

@inproceedings{bench-msds,
  title = {Multi-source Distributed System Data for {AI}-Powered Analytics},
  author = {Nedelkoski, Sasho and Bogatinovski, Jasmin and Mandapati, Ajay Kumar and Becker, Soeren and Cardoso, Jorge and Kao, Odej},
  booktitle = {Service-Oriented and Cloud Computing (ESOCC)},
  year = {2020},
  publisher = {Springer},
  doi = {10.1007/978-3-030-44769-4_13}
}

@book{m-box-jenkins,
  title = {Time Series Analysis: Forecasting and Control},
  author = {Box, George E. P. and Jenkins, Gwilym M.},
  publisher = {Holden-Day},
  year = {1970}
}

@inproceedings{m-anomaly-transformer,
  title = {Anomaly Transformer: Time Series Anomaly Detection with Association Discrepancy},
  author = {Xu, Jiehui and Wu, Haixu and Wang, Jianmin and Long, Mingsheng},
  booktitle = {ICLR},
  year = {2022}
}

@inproceedings{m-anomaly-bert,
  title = {AnomalyBERT: Self-Supervised Transformer for Time Series Anomaly Detection using Data Degradation Scheme},
  author = {Jeong, Yungi and Yang, Eunseop and Ryu, Jong Hwan and Park, Imseong and Kang, Myungjoo},
  booktitle = {ICLR Workshops},
  year = {2023}
}

@inproceedings{m-dcdetector,
  title = {DCdetector: Dual Attention Contrastive Representation Learning for Time Series Anomaly Detection},
  author = {Yang, Yiyuan and Zhang, Chaoli and Zhou, Tian and Wen, Qingsong and Sun, Liang},
  booktitle = {KDD},
  year = {2023}
}

@inproceedings{m-memto,
  title = {MEMTO: Memory-Guided Transformer for Multivariate Time Series Anomaly Detection},
  author = {Song, Junho and Kim, Keonwoo and Oh, Jeonglyul and Cho, Sungzoon},
  booktitle = {NeurIPS},
  year = {2023}
}

@inproceedings{m-sub-adjacent,
  title = {Sub-Adjacent Transformer: Improving Time Series Anomaly Detection via Sub-Adjacent Window Reconstruction},
  author = {Yue, Wenzhen and Ying, Xianghua and Guo, Ruohao and Chen, DongDong and Shi, Ji and Xing, Bowei and Zhu, Yuqing and Chen, Taiyan},
  booktitle = {IJCAI},
  year = {2024}
}

@inproceedings{m-times-net,
  title = {TimesNet: Temporal 2D-Variation Modeling for General Time Series Analysis},
  author = {Wu, Haixu and Hu, Tengge and Liu, Yong and Zhou, Hang and Wang, Jianmin and Long, Mingsheng},
  booktitle = {ICLR},
  year = {2023}
}

@inproceedings{m-itransformer,
  title = {{iTransformer}: Inverted Transformers Are Effective for Time Series Forecasting},
  author = {Liu, Yong and Hu, Tengge and Zhang, Haoran and Wu, Haixu and Wang, Shiyu and Ma, Lintao and Long, Mingsheng},
  booktitle = {ICLR},
  year = {2024}
}

@inproceedings{m-patchtst,
  title = {A Time Series is Worth 64 Words: Long-term Forecasting with Transformers},
  author = {Nie, Yuqi and Nguyen, Nam H and Sinthong, Phanwadee and Kalagnanam, Jayant},
  booktitle = {ICLR},
  year = {2023}
}

@inproceedings{m-moment,
  title = {{MOMENT}: A Family of Open Time-series Foundation Models},
  author = {Goswami, Mononito and Szafer, Konrad and Choudhry, Arjun and Cai, Yifu and Li, Shuo and Dubrawski, Artur},
  booktitle = {ICML},
  year = {2024}
}

@inproceedings{m-gpt4ts,
  title = {One Fits All: Power General Time Series Analysis by Pretrained {LM}},
  author = {Zhou, Tian and Niu, Peisong and Wang, Xue and Sun, Liang and Jin, Rong},
  booktitle = {NeurIPS},
  year = {2023}
}

@article{m-notears,
  title = {{DAGs} with {NO TEARS}: Continuous Optimization for Structure Learning},
  author = {Zheng, Xun and Aragam, Bryon and Ravikumar, Pradeep and Xing, Eric P},
  journal = {NeurIPS},
  year = {2018}
}

@article{m-tranad,
  title = {{TranAD}: Deep Transformer Networks for Anomaly Detection in Multivariate Time Series Data},
  author = {Tuli, Shreshth and Casale, Giuliano and Jennings, Nicholas R},
  journal = {VLDB},
  year = {2022}
}

@inproceedings{m-gdn,
  title = {Graph Neural Network-Based Anomaly Detection in Multivariate Time Series},
  author = {Deng, Ailin and Hooi, Bryan},
  booktitle = {AAAI},
  year = {2021}
}

@inproceedings{m-mtadgat,
  title = {Multivariate Time-Series Anomaly Detection via Graph Attention Network},
  author = {Zhao, Hang and Wang, Yujing and Duan, Juanyong and Huang, Cong and Cao, Defu and Tong, Yunhai and Xu, Bixiong and Bai, Jing and Tong, Jie and Zhang, Qi},
  booktitle = {ICDM},
  year = {2020}
}

@inproceedings{m-omnianomaly,
  title = {Robust Anomaly Detection for Multivariate Time Series through Stochastic Recurrent Neural Network},
  author = {Su, Ya and Zhao, Youjian and Niu, Chenhao and Liu, Rong and Sun, Wei and Pei, Dan},
  booktitle = {KDD},
  year = {2019}
}

@inproceedings{m-usad,
  title = {UnSupervised Anomaly Detection on Multivariate Time Series},
  author = {Audibert, Julien and Michiardi, Pietro and Guyard, Fr{\'e}d{\'e}ric and Marti, S{\'e}bastien and Zuluaga, Maria A.},
  booktitle = {KDD},
  year = {2020}
}

@article{m-vus,
  title = {Volume Under the Surface: A New Accuracy Evaluation Measure for Time-Series Anomaly Detection},
  author = {Paparrizos, John and Kang, Yuhao and Boniol, Paul and Tsay, Ruey S and Palpanas, Themis and Franklin, Michael J},
  journal = {NeurIPS Datasets \& Benchmarks},
  year = {2022}
}

@article{m-pa-f1-critique,
  title = {Towards a Rigorous Evaluation of Time-Series Anomaly Detection},
  author = {Kim, Siwon and Choi, Kukjin and Choi, Hyun-Soo and Lee, Byunghan and Yoon, Sungroh},
  journal = {AAAI},
  year = {2022}
}

@article{malhotra2016lstmae,
  title = {{LSTM}-based Encoder-Decoder for Multi-sensor Anomaly Detection},
  author = {Malhotra, Pankaj and Ramakrishnan, Anusha and Anand, Gaurangi and Vig, Lovekesh and Agarwal, Puneet and Shroff, Gautam},
  journal = {ICML Anomaly Detection Workshop},
  year = {2016}
}

\end{document}